\documentclass[letterpaper, 10 pt, conference]{ieeeconf}  
\usepackage[dvipsnames]{xcolor}
\usepackage{amsmath}
\usepackage{graphicx}
\usepackage{listings}
\usepackage{amssymb}

\newcommand\mdoubleplus{\mathbin{+\mkern-10mu+}}

\IEEEoverridecommandlockouts                              

\begin{document}

\title{\LARGE \bf
Symbol Guided Hindsight Priors for Reward Learning from Human Preferences
}

\author{Mudit Verma$^{1}$ and Katherine Metcalf$^{2}$
\thanks{$^{1}$SCAI, Arizona State University (work done while at Apple), AZ , 85281 {\tt\small muditverma@asu.edu}}
\thanks{$^{2}$Apple, Cupertino, CA, 95014, {\tt\small kmetcalf@apple.com}}
}
\maketitle
\thispagestyle{empty}
\pagestyle{empty}

\begin{abstract}
Specifying rewards for reinforcement learned (RL) agents is challenging. Preference-based RL (PbRL) mitigates these challenges by inferring a reward from feedback over sets of trajectories. However, the effectiveness of PbRL is limited by the amount of feedback needed to reliably recover the structure of the target reward. We present the PRIor Over Rewards (PRIOR) framework, which incorporates priors about the structure of the reward function and the preference feedback into the reward learning process. Imposing these priors as soft constraints on the reward learning objective reduces the amount of feedback required by half and improves overall reward recovery. Additionally, we demonstrate that using an abstract state space for the computation of the priors further improves the reward learning and the agent's performance.
\end{abstract}

\section{Introduction}

Recent works use binary preference feedback over pairs of trajectories to iteratively encode desired agent behaviors in a reward function and train the agent on the reward function via standard reinforcement learning (RL) \cite{pebble, surf, christiano, open-ai-langugae-pref}. Preference-based RL (PbRL) assumes a teacher provides preference feedback that maximizes a desired target reward function. The reward function is inferred from the preference labels by assigning rewards to states such that the sum of rewards in the preferred trajectory is larger than the sum of rewards in the dis-preferred trajectory. Initial works to integrate PbRL into Deep RL methods, like \cite{christiano}, suffered from low feedback efficiency, which was improved upon by PEBBLE \cite{pebble} and SURF \cite{surf}, however their key insights were to improve the quality of queries to the teacher and grow the size of the preference dataset. We hypothesize that some states play are more critical role in the teacher's preference feedback, those states can be inferred via auxiliary prediction processes, and information about those states can be incorporated into the reward objective via soft constraints. Therefore, we present PRIOR, which incorporates priors computed from the sampled trajectories and the preference-labeled trajectory pairs into the reward function updates, and show improvements in policy performance that are further increased when the priors are computed with respect to a human interpretable state-abstraction in the form of boolean propositions as symbols.



\section{Background}
The popular paradigm of learning from human preferences involves an agent acting in an environment $\mathcal{E}$ by sensing an observation $o_t \in \mathcal{O}$ at time $t$. The agent uses its policy $\pi$ to take an action $a_t \in \mathcal{A}$ to receive the next observation. As in traditional reinforcement learning, our agent models the environment as an MDP tuple $(\mathcal{O, T, A, R})$ where $\mathcal{T}$ is the transition function determined by the environment dynamics and $\mathcal{R}$ is the human in the loop's underlying reward function.

We follow the general learning from human preferences paradigm as in \cite{pebble, surf, christiano}  and infer the human target reward function from binary preference labels over pairs of trajectories.  Learning from human preferences via binary feedbacks on trajectory queries utilizes the Bradley Terry model to obtain probability estimates, $P_{\psi}$, of the human's preference by treating the predicted returns for each trajectory in a given pair, $\tau_0, \tau_1$ in dataset $\mathcal{D}$, as logits and applying softmax over the pair of predicted returns. The returns are computed by summing over the predicted reward for each state-action pair in each trajectory. These probability estimates are used to compute and subsequently minimize the cross-entropy between the predictions and the true human labels as, 
\begin{equation}
\begin{split}
    \mathcal{L}_{CE} = \underset{(\tau_0, \tau_1, y) \sim \mathcal{D}}{-\mathbb{E}}[
    & y(0)\log P_{\psi}[\tau_0 \succ \tau_1] \\ 
    & + y(1)\log P_{\psi}[\tau_1 \succ \tau_0]]
\end{split}
\end{equation}

In contrast to \cite{pebble, christiano, surf} we assume a mapping $M : \mathcal{O} \to \mathcal{S}$ that provides a many to one mapping from observations to symbolic abstractions. Such a mapping allows for a more robust computation of priors over the reward function as against to using the agent observations directly (Section \ref{symbol-abstraction} discusses more on interpretable symbolic abstraction).


We use the PEBBLE \cite{pebble} algorithm as the backbone for our experiments, however PRIOR's contributions are not limited to the PEBBLE algorithm. Since we propose priors operationalized as soft constraints to the learnt reward model $r_{\psi}$, we note that our method is complementary to existing methods that typically attempt to improve agent's performance or the reward recovery by improving the query \cite{pebble, surf} to the human in the loop.

\section{Method}
\begin{figure*}[h!]
  \centering
  \includegraphics[scale=0.28]{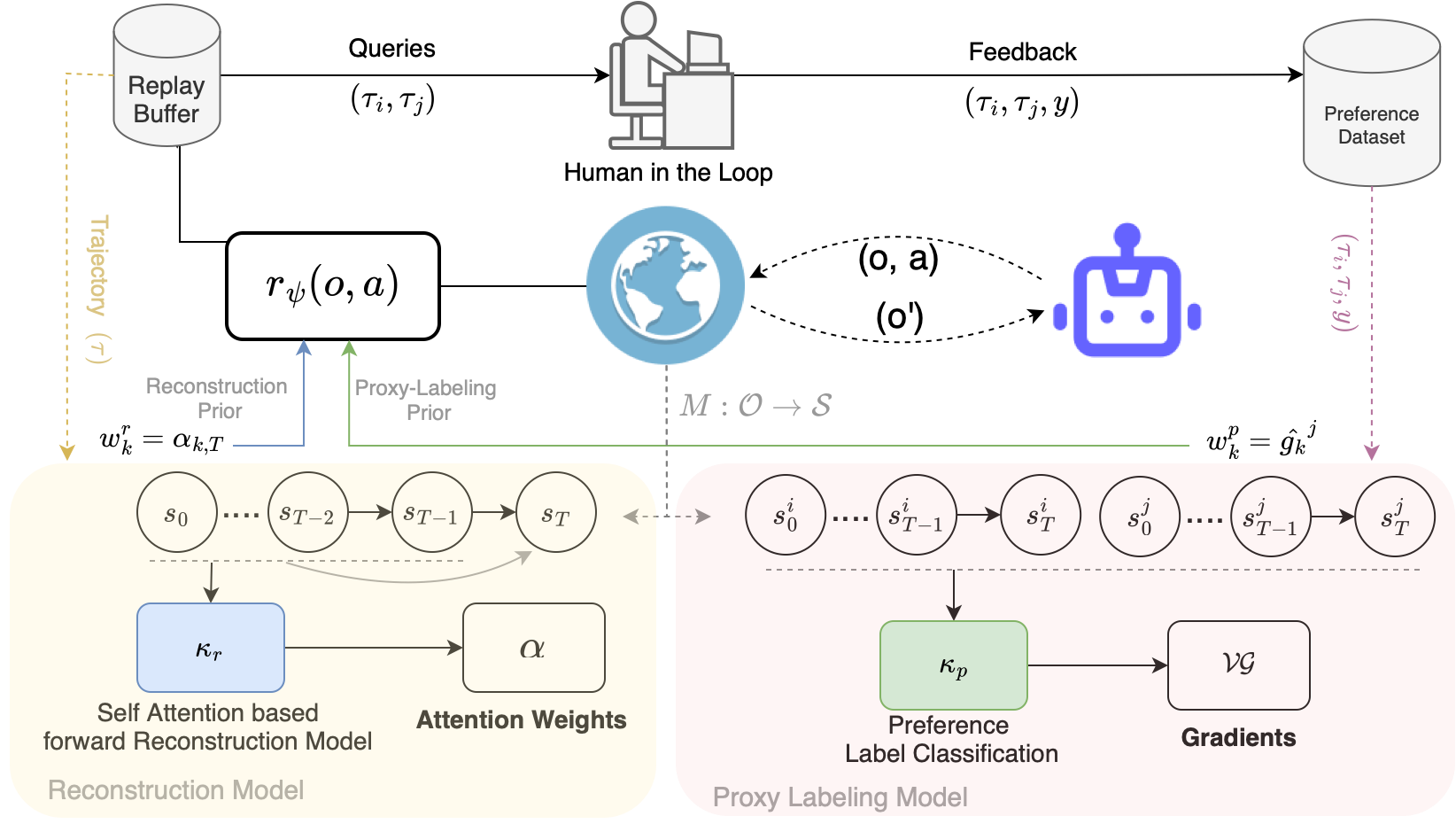}
  \caption{Overview of the PRIOR (PRIor Over Rewards) framework. PRIOR follows the typical preference based learning paradigm \cite{pebble,christiano} that uses trajectories stored in the agent's replay buffer to query the human in the loop for their binary feedback. The reward learning objective given in Eq. \ref{eq:reward-objective} is used to obtain the reward function $r_{\psi}(o,a)$. The loss objective utilizes a Reconstruction Prior, which computes a prior $w^r$ about the reward using attention weights derived from a forward-prediction world model called the Reconstruction Model, and a Proxy-labelling Prior that utilizes a self-attention based Preference label scheme to compute importance of the states in the pair of trajectories via Vanilla Gradients and uses that as the $w^p$ prior. In PRIOR the priors are computed over a state abstraction built using a given observation to symbol mapping function $M$, whereas in observation-PRIOR (another variant of PRIOR) we directly use observations $o_i$.}
  \label{fig:overview}
\end{figure*}

Our approach weights states during the reward function update according to their influence in determining the teacher's preference. First, we define important states as those with high state visitation frequency under a given policy \(\pi\) and environment dynamics. We use an auxiliary reconstruction task, \(\mathcal{R}: s_{t-k}...s_{t} \rightarrow s_{t+1}\), where the contribution of each state in window of size \(k\) correlates with state visitation, to compute a \textit{reconstruction prior} (Section \ref{sec:reconstruction_prior}). While, in the presence of a large number of states, count based methods can be used to compute state visitation \cite{count-exploration} which would still be approximate measures, instead we utilize the observation that for forward predicting world-models \cite{world-models} (Reconstruction Model, Figure \ref{fig:overview}), the attention over a state in the history when predicting a future state correlates with state co-occurrence frequency. The prior encourages the magnitude of each state's reward to have a mass proportional to the state's contribution to the future state prediction. 

To ensure that states are weighted appropriately, weights are derived from trajectory pairs so that states are not up-weighted in a trajectory that is preferred only because of a negative outcome in the dis-preferred trajectory. To address such scenarios, we introduce a \textit{proxy-labelling prior} (Section \ref{sec:proxy_labelling_prior}) to encourage state rewards in a pair of trajectories to correlate with their importance for predicting the correct preference label. The prior is computed using a classification model that is learnt to predict preference labels given both trajectories as input, unlike the reward function which only has access to individual states. 

Finally, we compute both priors over an abstract state representation interpretable to the human-teacher to further boost the agent's reward recovery performance. Humans use symbolic structures when selecting preferences \cite{blackbox, lingua-franca} and using the vocabulary of such a symbolic space improves reward attribution to states. The priors use the given symbolic vocabulary and are incorporated into reward function learning as soft constraints \cite{soft-constraints} on the reward-learning objective as detailed in the next sections. Figure \ref{fig:overview} shows an overview of the PRIOR framework.

\begin{figure*}[h!]
  \centering
  \includegraphics[scale=0.23]{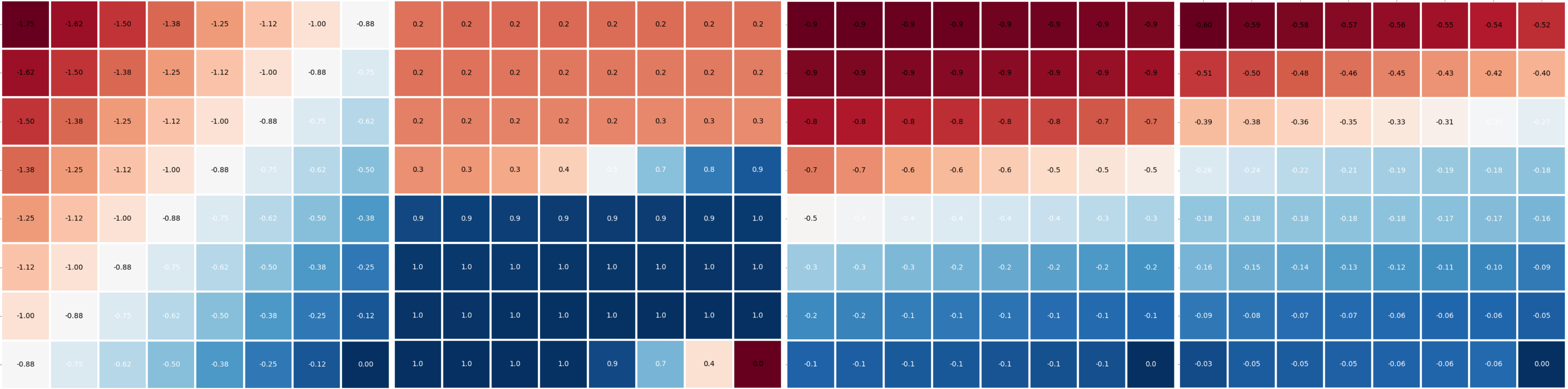}
  \caption{Visualizations of the learned reward function for each state in the 8x8 gridworld. The agent always starts at the cell in the upper left and should move as quickly as possible to the cell in the lower right. A well-recovered reward function is negative everywhere and assigns smaller rewards the further a state is from the goal location. From left to right the reward functions are the teacher's (underlying) target, PEBBLE's, PRIOR's with observations, and PRIOR's with symbols. The value in each cell of a gridworld indicates the maximum reward that can be obtained by an action when the agent is at that cell. The PEBBLE-learned reward function assigns smaller rewards to cells further from the goal, but it assigns positive rewards everywhere, whereas PRIOR achieves both negative rewards and smaller rewards further from the goal.}
  \label{fig:combined-prior-v-pebble}
\end{figure*}

\subsection{Reconstruction Prior} \label{sec:reconstruction_prior}
The reconstruction model (\(\mathcal{R}\)) is a transformer \cite{transformer} with a single self-attention layer \cite{single-transformer} and head to reconstruct state \(s_{t+1}\) given the \(k\) prior states \(\tau_{[t-k:t]}\). Once learned, the reconstruction model's attention weights encode the ``importance'' of a state \(s_i\) (\(t-k < i < t\)) in a trajectory \(\tau\). The reconstruction prior \(w_i^r\) imposes the importance of a state \(s_i\) to be proportional to its attention weight when predicting \(s_{t+1}\) from \(\tau_{[t-k:k]}\) by setting \(w_i^r = \alpha_{i}\), where \(\alpha_{i}\) is the attention weight for \(s_{i}\).
%
%
%
%
%
Finally, the reconstruction prior is incorporated into reward function ($r_{\psi}(s,a) \to [-1,1]$) learning with the following auxiliary loss function, \(L_r(\tau, r_\psi) = D_{KL}(w_{i}^{r}(\tau) \mathbin\Vert \sigma(r_\psi(\tau)))\), where \(w_{i}^{r}(\tau)\) is the reconstruction prior over states in \(\tau\) and $\sigma$ is the softmax function. In this work, the trajectory's final state is used as \(s_{t+1}\) to compute the relative importance of each state \(s_{i}\) and \(k = |\tau| - 1\).

\subsection{Proxy Labelling Prior}\label{sec:proxy_labelling_prior}
The proxy labelling model, $\mathcal{M_P}$, which is a self-attention sequence classifier based on the Transformer Architecture which takes in the two sequences of trajectories: (1) each step is embedded using a state-action encoder and (2) each trajectory is embedded using the self-attention model. The trajectory embeddings are fed to a fully connected layer that predicts the binary preference.
The proxy labelling model assigns credit to the states that contribute to a correct preference prediction given a pair of trajectories $(\tau^0, \tau^1)$. Each state's contribution is computed using Explanations as VanillaGradient \cite{vanillagrad} for trajectory pair $\tau_j; j \in \{0,1\}$ as: \\ 
\vspace{-0.5cm}
\begin{multline}
    g_i^j = \mathcal{VG}(\hat{y_i}, I_{s_i^j, a_i^j}, \mathcal{M_P}) \quad ; i \in \{1,2..T_j\} \\
    \hat{g_k}^{j = 1-\hat{y}} = \begin{cases}
    \min_{x}g_x^j - g_k^j & \text{;}g_k^j >0\\[0.5pt]
    g_k^j &\text{;} g_k^j \leq 0 \\[0.5pt] 
    \end{cases} 
\vspace{-0.8cm}
\end{multline}
where $\mathcal{VG}(y, \mathcal{I}, \mathcal{M})$ computes the vanilla-gradient of the output $\hat{y}$ with respect to input $\mathcal{I}$ for a deep neural network $\mathcal{M}$. We use the $\hat{y}$ as predicted by the Proxy-Labelling Model and compute the gradients $g_i^j$ for both the trajectories while ensuring that the gradients of the dis-preferred trajectory $1-\hat{y}$ are always negative ($\hat{g_k}^{j = 1-\hat{y}}$). This is to impose the condition that rewards of the preferred trajectory should be higher than that of the dis-preferred trajectory. Finally, a loss between the softmax gradients of the two trajectories (concatenation shown by $\mdoubleplus$) and the softmax reward gives the $L_p$ loss function: 
\begin{equation}
    L_p(\tau^0, \tau^1, r_{\psi}) = D_{KL}(\sigma(r_\psi^{\tau^{\hat{y}}} \mdoubleplus r_\psi^{\tau^{1-\hat{y}}})) \mathbin\Vert \sigma(g^{\hat{y}} \mdoubleplus \hat{g}^{1-\hat(y)}) )
\end{equation}

\subsection{Combined Losses \& Algorithm}
For some positive scalar coefficients $\lambda_p, \lambda_{r0}, \lambda_{r1}$, we take a linear combination of the three losses as follows: 
\begin{equation} \label{eq:reward-objective}
\begin{split}
\mathcal{L}_{reward}(\tau_0, \tau_1, r_\psi) & = \mathcal{L_{CE}}(\tau_0, \tau_1, r_\psi) + \lambda_p\mathcal{L}_p(\tau_0, \tau_1, r_\psi) +\\ 
 & \lambda_{r0}\mathcal{L}_r(\tau_0, r_\psi) + \lambda_{r1}\mathcal{L}_r(\tau_1, r_\psi)
\end{split}
\end{equation}
We use a similar training loop to \cite{pebble,christiano}, where we use our reward-learning objective. In addition, we train the reconstruction model along with updates to the agent's policy (which can also be done as a parallel task) followed by training the surrogate proxy-labeling model for a given number of epochs, each time new feedback is provided.

\subsection{Interpretable Symbol Abstraction} 
\label{symbol-abstraction}
Rather than using the observation space $\mathcal{O}$ used by the agent for computing $L_r, L_p$, we instead view the trajectories in terms of a vocabulary of binary predicates $\mathcal{S} = \{s_k\} \; k = \{1..K\}$.  Hence we assume a mapping function $M : \mathcal{O} \to \mathcal{P}(\mathcal{S})$, where $\mathcal{P}(x)$ represents the power set of $x$. The mapping function can be elicited from the human in the loop or extracted from domain descriptions as described in \cite{blackbox}. Once the mapping function is available, PRIOR can use the symbols to better interpret the meaning of the preference labels, because the labels are given by the human who tends to accept explanations and reason in symbolic terms \cite{lingua-franca} and thus priors on the reward values over states built over the symbolic abstraction becomes very useful especially when such priors are being used as soft-constraints.

\vspace{-0.2cm}
\section{Results}

We present a preliminary investigation of PRIOR on a Gridworld, where the teacher's preference is for the agent to reach the lower right corner starting from the top left corner. Gridworlds allow us to hand-define a reward function with a known structure, making it easier to visualize, analyze, and compare the structure of the learned reward functions. We use a synthetic, perfect teacher as in \cite{pebble} for binary feedbacks based on the negative Manhattan distance between the agent and the goal; trajectories with lower mean distance to the goal are preferred. QTable \cite{qtable} is used to learn all Gridworld policies (\(\epsilon = 0.5\); \(\alpha = 0.1\); \(\gamma = 0.99\)). We use the following symbols for the gridworld, 
\texttt{at\_top\_edge,  at\_left\_edge, at\_right\_edge, at\_bottom\_edge, at\_top\_left\_corner, at\_top\_right\_corner, at\_bottom\_left\_corner, at\_bottom\_right\_corner.}
The boolean symbols evaluate to True when the agent is present in the corresponding location (for example, the left edge or the top right corner).

We evaluate reward function quality by assessing how well the learned reward recovers the structure of the oracle reward and whether the learned reward generates a goal-reaching policy. For the learned reward to recover the structure of the oracle's reward, it must be negative everywhere, and rewards further from the goal must be smaller than rewards closer to the goal. For situations where the learned rewards are always negative, we use Episode Return Correlation Distance (EPC, \textit{lower} values imply \textit{higher} similarity)\cite{epc} between the ground truth and learned reward to assess how well the reward function learned to assign smaller rewards further from the goal. 
Note that EPC is not sensitive to whether the compared rewards functions are negative versus positive. Below, we compare PEBBLE \cite{pebble}, PRIOR with observations (O-PRIOR), and PRIOR with symbols to the ground truth reward. All experiments reported are run on an 8x8 Gridworld and we found these results generalize to a larger grid sizes. Unless specified otherwise, results are reported for 40 queries of length 8 steps. The robustness of the reward learning methods is evaluated according to the number of queries and the query length required to recover the reward function and a goal-reaching policy. Additionally, we evaluate the quality of the recovered reward structure when the reward function is forced negative, $r = \frac{\tanh(x)-1}{2}$, to ensure a goal-reaching policy (the goal has reward 0).

\subsection{Number of Queries}
PRIOR obtains the desired reward function and a goal-reaching policy with fewer queries (40 queries) than PEBBLE (96 queries). Fig. \ref{fig:combined-prior-v-pebble} visualizes (left to right) the ground truth, PEBBLE, O-PRIOR, and PRIOR reward functions shows that PEBBLE struggles to get required reward-recovery within 40 queries (and infact takes over 96 queries to obtain the required performance).
    
\subsection{Query Length}
For shorter queries, \(\{3, 5\}\), both PRIOR and PEBBLE were unable to recover the environment reward and goal-reaching policies, whereas for long queries, \({8, 12}\), PRIOR learns rewards that are negative everywhere and decrease further from the goal while PEBBLE was unable to learn negative rewards everywhere.
    
\subsection{Forced Negative} 
We suspected that our baseline may be struggling with getting the first aspect of reward-recovery right, that is to produce negative rewards everywhere. We investigated the performance of PRIOR versus PEBBLE by constraining the reward prediction to always lie in the interval $[-1,0]$ in an attempt to ablate and check the second aspect of reward-recovery under concern, rewards farther from goal are less. We find that PRIOR better recovered the ground truth reward function's structure (EPC=0.54, moderate recovery) than PEBBLE (EPC=0.82, comparatively poor recovery), where both PRIOR and PEBBLE were able to obtain the optimal policy (as intended by imposing the constraint).
    

Finally, we ablate PRIOR variants and find that even though O-PRIOR can get a goal-reaching policy, the reward recovery is worse than PRIOR.

\section{Discussion}
We presented PRIOR, a method to incorporate prior knowledge about the structure of rewards functions into the reward learning process, and demonstrated that computing priors with respect to symbols instead of observations further improves learned reward quality. Our results show, on a Gridworld task, PRIOR is better able to recover rewards that are negative everywhere and decrease further from the goal then the previous state of the PEBBLE \cite{pebble}. Expanding upon our promising results, we found similar performance benefits of PRIOR over PEBBLE on a variant of Montezuma's Revenge Level 1 \cite{blackbox} and our future work includes extensive evaluation of PRIOR on more complex domains and preferences.


\end{document}